\documentclass{article}

\usepackage{microtype}
\usepackage{graphicx}
\usepackage{amsmath,amssymb}
\usepackage{subfigure}
\usepackage{booktabs} 

\usepackage{hyperref}

\usepackage[accepted]{icml2020}

\icmltitlerunning{Am I Building a White Box Agent or Interpreting a Black Box Agent?}

\begin{document}

\twocolumn[
\icmltitle{Am I Building a White Box Agent or Interpreting a Black Box Agent?}



\icmlsetsymbol{equal}{*}

\begin{icmlauthorlist}
\icmlauthor{Tom Bewley}{bris}
\end{icmlauthorlist}

\icmlaffiliation{bris}{Department of Engineering Mathematics, University of Bristol, Bristol, United Kingdom}

\icmlcorrespondingauthor{Tom Bewley}{tom.bewley@bristol.ac.uk}

\icmlkeywords{Machine Learning, ICML}

\vskip 0.3in
]



\printAffiliationsAndNotice{}  

\begin{abstract}
The rule extraction literature contains the notion of a fidelity-accuracy dilemma: when building an interpretable model of a black box function, optimising for fidelity is likely to reduce performance on the underlying task, and vice versa. I reassert the relevance of this dilemma for the modern field of explainable artificial intelligence, and highlight how it is compounded when the black box is an agent interacting with a dynamic environment. I then discuss two independent research directions -- building white box agents and interpreting black box agents -- which are both coherent and worthy of attention, but must not be conflated by researchers embarking on projects in the domain of agent interpretability.
\end{abstract}

\section{Introduction}

The rule extraction literature, most vibrant in the late 1990s and early 2000s, was in many ways the ideological precursor to the modern field explainable artificial intelligence (XAI). Two decades ago, state-of-the-art machine learning models were modestly-sized by today's standards, but there was nonetheless an understanding that impactful real-world applications, particularly those that are safety-critical, demand a depth of scrutiny and assurance that is unattainable with opaque \textit{black boxes}. The objective of rule extraction was to approximate trained machine learning models, most commonly artificial neural networks, with propositional rule structures that can be more readily understood and analysed. In those times as in these, there was great interest in the instrumental goal of defining quantitative performance measures for the extraction process, which provide some agreed-upon standard for judging candidate approaches.

Amongst the most influential proposals from this period is the FACC framework \cite{andrews1995survey}, which outlines four criteria for evaluating rule extraction:

\paragraph{Fidelity} The degree to which the rule set's outputs agree with those of the black box model.
\paragraph{Accuracy} The ability of the extracted rule set to correctly predict outputs for unseen samples; its performance on the underlying task.
\paragraph{Consistency} The stability of extracted rules over reasonable variations in the training conditions, such as using different subsets of a training dataset.
\paragraph{Comprehensibility} The ease with which the rule set can be read and understood by a human end-user, which might be quantified in terms of the number of rules, or the number of antecedents per rule. 

Several years after the introduction of FACC, Zhou published a highly-perceptive critique of this and similar evaluation frameworks \cite{zhou2004rule}. While he takes no issue with any of the four criteria in isolation, he identifies a fundamental conflict between the first two: fidelity and accuracy. Unless the black box represents a perfect and unique solution to the underlying task, a rule set with zero accuracy loss will have nonzero fidelity loss, and vice versa. Furthermore, if either loss is nonconvex, there may be solutions with very high performance on one metric, but very low performance on the other. Zhou argues that a compromise is never truly what is needed for any realistic application, and that either fidelity or accuracy should be chosen at the exclusion of the other. 

This point has been partly internalised by XAI researchers in the years since, but many continue to evaluate their models through an ad hoc combination of metrics, some of which are in conflict. I suggest that this is a significant source of muddled thinking in the XAI field, and may indeed be hampering progress.

My particular interest is in the analysis and interpretation of black box agents, such as those trained by reinforcement learning (RL), rather than supervised classification or regression models. This context boasts the added complexities of non-i.i.d. data (due to the presence of closed-loop interaction) and nontrivial credit assignment (via possibly-delayed rewards). Here, the force of Zhou's point is not only preserved but amplified. There are many more ways in which an agent model may be evaluated, all of which appear similar at first glance but are mutually-contradictory in the details. Several recent papers \cite{bastani2018verifiable,coppens2019distilling,nageshrao2019interpretable,bewley2020modelling} do not fully disambiguate the various objectives, and as a result, appear to leave both author and reader alike somewhat conflicted as to what the purpose of the whole exercise has been. Going forward, I propose that researchers interested in the general problem of interpretable agents should frequently ask themselves the following question, which this paper seeks to clarify:
\begin{center}
\textit{``Am I building a white box agent\\ or interpreting a black box agent?"}
\end{center}

\section{Summary of Zhou's Argument}

Consider the supervised learning problem of mapping input vectors $\textbf{x}\in\mathcal{X}$ to discrete or continuous action labels $y\in\mathcal{Y}$. We can frame this problem as that of approximating a latent function $h$ that performs the mapping exactly: $y=h(\textbf{x})$. In the domain of rule extraction (and likewise in XAI), our typical starting point is to be given a trained black box function $f_B$ which goes some way to approximating $h$ but is almost certaintly not optimal\footnote{If we knew that $f_b$ perfectly matched $h$, we may have significantly less of an incentive to explain it in the first place.}. Using our favourite method, we create a \textit{white box} approximation of $f_B$, denoted by $f_W$ (this might be a rule set, but could equally be a linear regressor, Na\"{\i}ve Bayes classifier or any other model class that we deem interpretable). For a given loss function $\mathcal{Y}\times\mathcal{Y}\rightarrow[0,1]$, the fidelity and accuracy of $f_W$ can be measured on a test dataset $X$:
\begin{equation}
\text{fidelity}_{W}=1-\mathbb{E}_{\textbf{x}\in X}\ \ell(f_W(\textbf{x}),f_B(\textbf{x}));\\
\end{equation}
\begin{equation}
\text{accuracy}_{W}=1-\mathbb{E}_{\textbf{x}\in X}\ \ell(f_W(\textbf{x}),h(\textbf{x})).\\
\end{equation}
Given a particular sample $\textbf{x}^*$ for which $\ell(f_B(\textbf{x}^*),h(\textbf{x}^*))>0$ (i.e. the black box's prediction has some error), we have the choice of either updating $f_W$ to match $f_B$ on this sample, or updating it to match $h$. The first approach would improve $\text{fidelity}_{W}$ at the expense of $\text{accuracy}_{W}$, and the second would improve $\text{accuracy}_{W}$ at the expense of $\text{fidelity}_{W}$. Our inability to improve both metrics in general is the \textit{fidelity-accuracy dilemma}.

Zhou identifies two valid approaches to resolving the dilemma. I retain his original names for these, noting that they can be updated for the XAI era by reading ``rule extraction" as ``white box induction" or ``explainability", and generalised by reading ``neural networks" as ``black boxes":
 
\paragraph{Rule extraction using neural networks} Here the goal is to obtain accurate and interpretable white box models, and as such, the key metric is \textbf{accuracy}. Black boxes are used to guide learning, either by assisting with data augmentation or providing `soft labels' that convey additional information about uncertainties. Ultimately, the aim is to replace them entirely, so they have the role of stepping stones to the interpretable models we really want, rather than the stars of the show in themselves. As Zhou notes, several researchers have found that extracted rule sets can generalise better than the original neural networks when tested on unseen samples \cite{craven1995extracting,setiono2000extracting,zhou2003extracting}. In such cases, pursuing increased fidelity would be harmful to the ultimate objective.

\paragraph{Rule extraction for neural networks} Here the goal is to understand the working mechanisms of trained black box models, by finding interpretable approximations that replicate their behaviour as faithfully as possible. We are now interested in optimising for \textbf{fidelity}, and accuracy on the underlying task is an irrelevant and misleading metric since we ultimately intend to continue using the black box in deployment. In fact, if the white box generalises better than the black box this is an actively bad result, because it may obscure our understanding of the black box's limitations.

It is Zhou's view, and ours alike, that there is no coherent halfway point between these two objectives. Either we want to replace the black box with a white box, in which case we optimise for accuracy, or we want to retain the black box and use the white box to understand it, so we optimise for fidelity. A great deal of confusion and miscommunication could be avoided by splitting the popular FACC framework into (1) ACC for the former case, and (2) FCC for the latter.

\section{The Agent-based Context}

Zhou's argument refers exclusively to the supervised learning domain. It sentiment remains valid in the dynamic agent-based context, but the situation also becomes significantly more complicated. In this discussion, we can adopt the formalism of a Markov Decision Process (MDP). 

At each timestep $t$ in an MDP, a dynamical system referred to as an \textit{environment} has a state $S_t=s\in\mathcal{S}$, which is perceived by an agent and used to condition the sampling of its action $A_t=a\in\mathcal{A}$ from a policy function $\pi(a\vert s)$. The state for the following timestep, $S_{t+1}=s'$, and a scalar \textit{reward} value $R_{t+1}=r$, are sampled from a dynamics function $p(s',r\vert s,a)$ and the process continues iteratively until a termination condition is met (e.g. a time limit $t=T$ or one of a set of terminal states). A complete sequence of states, actions and rewards from initialisation to termination is called an \textit{episode}. 

Of particular interest for many problems in the MDP context is the \textit{return} $G(S_t)$, defined as the sum of reward obtained after visiting $S_t$, discounted by a factor $\gamma\in[0,1]$: $G(S_t)=\sum_{k=1}^\text{termination}\gamma^kR_{t+k}$. The goal of a reinforcement learning (RL) agent is to use exploration to find a policy $\pi$ that maximises the expected return, referred to as the \textit{value}, $V_\pi(s)=\mathbb{E}_\pi[G(s)]$, across the full distribution of states encountered when following that same policy. 

Situating ourselves once more at the starting point of the XAI investigation, we assume the existence of some black box policy $\pi_B$, which may or may not have been induced by RL, and a class of white box models $\Pi$. A popular choice for $\Pi$ is the class of binary, axis-aligned decision trees \cite{bastani2018verifiable,bewley2020modelling}. Remaining for the moment entirely agnostic about the data and learning algorithms used, our objective is to find the white box model $\pi_W\in\Pi$ that optimises... what, exactly?

What is it that we truly wish to achieve by our creation of the white box model $\pi_W$, and how does this relate to the existing black box $\pi_B$? I believe that the answer to this question, frequently bypassed or deemed too obvious to ask explicitly, is completely application-dependent.

A first point to note here is that there is no unambiguous notion of accuracy in an MDP. There are no ground truth ``correct" actions for an agent to take, and task performance is measured at a higher level of abstraction in terms of return and value. The FACC framework can therefore not be ported over directly for critique. The closest equivalent would be to replace the accuracy criterion with the expected value $\bar{V}_W=\mathbb{E}_{s\sim d_{W}} V_{\pi_{W}}(s)$, where $d_W:\mathcal{S}\rightarrow[0,1]$ is the distribution of states encountered by the white box when deployed as a closed-loop control policy in the MDP. This quantifies how well the white box is able to solve the underlying task and obtain high reward. The FACC acronym can be updated to read FVCC.

The \textbf{expected value} $\bar{V}_W$ may indeed be the metric we wish to optimise, but it should be clear by now that doing so gives absolutely no guarantee of fidelity to the black box. $\pi_B$ may be a very poor policy, which obtains low reward in the MDP and frequently exhibits calamitous failure. A high-performing $\pi_W$ would differ from such a $\pi_B$ in every sense worth talking about. An alternative problem we might pose, then, is to optimise for the \textbf{similarity of value} between the two policies, which we can denote by $|\bar{V}_W-\bar{V}_B|$. This is a completely valid objective, but also does not imply good fidelity. Expected value is generally highly nonconvex in the policy itself (this is what makes RL so difficult), meaning there are many ways to obtain identical overall task performance with radically divergent policies. 

Notice that here the possible divergence is twofold. The actions taken by $\pi_B$ and $\pi_W$ in any given state may differ in themselves, but this difference also results in the MDP transitioning to different states for the following timestep, which gradually leads the two policies to move into separate regions of the state space. This compounding divergence between two MDP policies when deployed is called \textit{covariate shift}, and is a primary source of difficulty in imitation learning \cite{ross2011reduction}. Of course, such discrepancies need not concern us if our objective is solely to create a high-value white box, or to match the aggregate value of the black box, but any shred of doubt about this issue is an indication that we may have chosen the wrong objective.

The notion of fidelity may initially seem easier to define unambiguously, but closer inspection shows there are just as many complexities. The key question is: over what dataset should we calculate a fidelity score? The most obvious choice is to use samples from the \textbf{state distribution encountered by the black box}, $d_B$, which gives us one measure of fidelity $F_B$ that is closest to a conventional supervised learning loss. Alternatively, we could use the \textbf{white box's own state distribution} $d_W$, which differs from $d_B$ due to covariate shift, and gives us a different fidelity measure $F_W$. Another option is to evaluate only on selected \textbf{states that we deem to be interesting} in some way, perhaps due to their critical influence on expected return, or because they are relevant to a particular question about the black box's behaviour. This gives a third fidelity measure, $F_{\text{interesting}}$, which once again will give different results in general. As tempting as it may be, we should not attempt to evaluate a white box agent trained for any variant of fidelity by measuring its expected value $\bar{V}_W$; this is simply the wrong perspective to take. Some high-fidelity white boxes may achieve high expected value, others may achieve low value, and optimising for fidelity alone provides no mechanism for disambiguating between the two.

A final and distinct metric worth considering is the \textbf{difference $\Delta$ between the state-action occupancy measures} of the two policies $\pi_W$ and $\pi_B$, which are the distributions of state-action pairs (not just states) encountered by the policies when deployed in the MDP. The white box's occupancy is $\rho_{W}:\mathcal{S}\times\mathcal{A}\rightarrow[0,1]$, and $\rho_{B}$ is similarly defined for the black box. Minimising $\Delta(\rho_{W},\rho_{B})$, where $\Delta$ is the Jensen-Shannon divergence, is the objective in \textit{generative adversarial imitation learning} \cite{ho2016generative}. It can be shown that there is a unique policy for any given occupancy measure $\rho$, so any $\pi_W$ that is perfectly optimised for occupancy will also obtain a perfect score on all measures of fidelity, and the same value as $\pi_B$, because it will be identical to that policy. However, whenever the available policy class $\Pi$ is heavily constrained for interpretability, and optimisation power is limited by the algorithmic tools available, our search is realistically for a local optimum, and improvements towards this point will not monotonically improve either fidelity or expected value\footnote{There are also various similarity measures $\Delta$ that could be chosen besides the Jensen-Shannon divergence, which would give different local optima.}.

Given all of these reasonable evaluation options, the FVCC acronym expands to become rather unwieldy, and almost certainly includes terms beyond those discussed here:
$$F_B,F_W,F_{\text{interesting}},...,\bar{V}_{W},|\bar{V}_{W}-\bar{V}_{B}|,\Delta(d_{W},d_{B})...,C,C$$
An extension of the FACC philosophy, which would have us attempt a compromise between all of these objectives, would result in a gruesomely complicated loss function, and an optimised $\pi_W$ that is likely to be somewhat unsatisfactory from all perspectives. A common approach taken in the literature is to pick one of the fidelity- or value-based objectives, train $\pi_W$ to optimise that metric alone, but then proceed to evaluate on one or more different metrics. 

For instance, in \cite{coppens2019distilling} a fuzzy decision tree model is trained to minimise $F_B$ with respect to a black box RL model in a Mario game emulator, but is then evaluated by measuring the average reward it obtains when deployed. As the tree grows larger, this metric stabilises at a value around $30\%$ less than the black box, with no improvement when the depth increases from $6$ to $7$. The fact that no performance gain is obtained despite the model doubling in size (and thus becoming significantly less interpretable) clearly illustrates the difference between optimising for fidelity and value. In \cite{bewley2020modelling}, CART decision trees are optimised for $F_B$ with a selection of car driving policies, but then evaluated in terms of the frequency of crashes (a rough proxy for $\bar{V}_W$). Medium-sized trees are around twice as effective at avoiding crashes compared with the largest ones, despite the latter having upwards of $99.9\%$ accuracy on an unseen test set; more evidence that the two metrics do not correspond. In \cite{bastani2018verifiable}, an interactive training method is followed to optimise a decision tree for $F_W$, and this is shown to indirectly improve $\bar{V}_W$ compared with optimising for $F_B$, but the authors then analyse the tree in order to prove stability and robustness properties of the black box itself. The validity of these analyses is debatable if the data distribution seen by the tree during training ($d_W$) is not representative of that actually encountered by the black box ($d_B$). And in \cite{nageshrao2019interpretable}, a car following controller is approximated with a Takagi-Sugeno rule set by optimisation of $F_B$, but the two policies are compared through the visual similarity in their outputted actions at each point in \textit{time}. This is a misleading metric, different from the similarity of outputs in \textit{state}, since the two policies will inevitably diverge to some extent in state space as a consequence of covariate shift.

By not being fully clear about the ultimate purpose of creating $\pi_W$, the above works risk misinterpreting its statistical relationship to $\pi_B$, and coming to false conclusions about the quality and validity of the proposed method. They may benefit from being unambiguously committed to a single research direction and optimisation problem. In the following section I briefly discuss two such directions, and the methods most suited to pursuing them.

\section{Two Alternative Research Directions}

\subsection{Building White Box Agents}

This direction approximately corresponds to ``rule extraction using neural networks" from Zhou's original paper. Here, the goal is to obtain an interpretable white box policy $\pi_W$ with strong performance in the underlying MDP environment as measured by the expected value $\bar{V}_W$. While RL is the most explicit approach for finding high-value policies, it is notoriously sample-inefficient and commonly requires tens of millions of samples to solve reasonable-sized problems \cite{mnih2015human}. In addition, some interpretable models, such as decision trees, are poorly suited to RL\footnote{Some notable efforts to rectify this shortcoming include \cite{pyeatt2003reinforcement}, \cite{roth2019conservative} and \cite{silva2020optimization}.} due to their lack of a differentiable loss function \cite{silva2020optimization}. For these reasons, a pre-existing black box policy $\pi_B$ may serve as a \textbf{useful stepping stone} for guiding the induction of $\pi_W$, especially during the earliest stages. This is the mentality of imitation learning.

The widespread effectiveness of the \textsc{DAgger} algorithm \cite{ross2011reduction} testifies to the value of interactive, online learning in this context. Rounds of learning may be mixed with periods of deployment of $\pi_W$, during which we can identify the current performance limitations of this policy, and use the knowledge gained to inform targeted querying of $\pi_B$ where it would be most useful.

Such selectivity with respect to the data obtained from the black box is likely to skew the white box away from being an exact replica of its behaviour, but this is no cause for concern, since researchers pursuing the direction of building white box agents should be passionately disinterested in measures of fidelity for its own sake. They should also recognise the auxiliary status of the black box, which will have served its purpose once the white box has attained a suitable level of performance, and may even be permanently discarded at this point. For high-stakes decisions, the complete removal of black box functionality may be precisely the result we are aiming for \cite{rudin2019stop}.

\subsection{Interpreting Black Box Agents}

This direction corresponds to Zhou's ``rule extraction for neural networks". The black box policy $\pi_B$ is placed front-and-centre in this analysis, which aims to accurately distil its behaviour into an interpretable representation. The extracted white box model, which we can call $\pi_W$ for consistency with the rest of this paper, should be viewed less as a policy in its own right, and more as as a \textbf{scientific instrument} with which to scrutinise $\pi_B$, explain its actions, and reveal its flaws, biases and sources of instability. Where uncertainties are present in the white box approximation, these should be made explicit rather than hidden or resolved away.

In this direction, it simply does not occur to us to try deploying $\pi_W$ as a policy in the underlying MDP, just as we would never attempt to win a soccer game by fielding a piece of sports analytics software. Its role is as a companion to $\pi_B$, providing a reason to trust the black box when it is deployed. When optimising for this direction, some variant of fidelity is the correct measure of quality. $F_B$ is the obvious default, but other options would be reasonable, depending on which practical questions we wish to answer about $\pi_B$ and which regions of the state space we are most interested in.

This direction also leaves room for different kinds of interpretable model, trained to predict something other than the black box agent's single next action given the current state. A good example of this is the Linear Model U-tree \cite{liu2018toward}, which attempts to approximate the policy's value function, as well as the environment's transition probabilities, in addition to the policy itself. I am particularly interested in the prospect of entirely new interpretable architectures, specialised for the task of modelling the dynamic interactions of black box agents with their environments, and freed from the constraint of being unnecessarily evaluated as policies in their own right.

\section{Conclusion}

I have sought to direct the attention of contemporary XAI researchers towards Zhou's fidelity-accuracy dilemma, and highlight how the dilemma is compounded by the additional complexities of a dynamic, agent-based context. When creating white box approximations of black box agent policies, a lack of clarity about the ultimate objective brings a risk of confused evaluation, and a misinterpretation of the statistical relationship between the black and white boxes. The two research directions of building white box agents and interpreting black box agents are both coherent and worthy of sustained effort, but must not be conflated since their optimisation problems are different. Before embarking on projects in the domain of agent interpretability, researchers should ask themselves which of the two problems they ultimately wish to solve, and remain committed to their answers.

\bibliography{fidelity_perf}
\bibliographystyle{icml2020}

\end{document}